\documentclass[11pt,a4paper]{article}
\usepackage[hyperref]{acl2020}
\usepackage{times}
\usepackage{latexsym}
\usepackage{multirow}
\usepackage{amsmath}
\usepackage{hhline}
\usepackage{t1enc}    
\usepackage{hyperref}
\usepackage{url}
\usepackage{bbm}
\usepackage{hyperref}
\usepackage{amssymb}
\usepackage[utf8]{inputenc}
\usepackage{amsmath}
\usepackage[para]{footmisc}
\usepackage{graphicx}
\usepackage{booktabs}
\usepackage{tabularx}
\usepackage{color}
\newcolumntype{L}{>{\raggedright\arraybackslash}X}
\usepackage{amssymb}
\usepackage{upquote}
\usepackage{multirow}
\usepackage{xfrac}
\usepackage{contour}
\usepackage[normalem]{ulem}
\usepackage{makecell}
\usepackage{arydshln}
\usepackage{wrapfig}
\usepackage{hhline}

\usepackage{microtype}

\aclfinalcopy 

\definecolor{prefix}{rgb}{0, 0, 0}
\definecolor{knob}  {rgb}{1, 0, 0}
\definecolor{bw}    {rgb}{1, 0, 0}
\definecolor{rw}    {rgb}{.7, .3, .3}
\definecolor{knob4}  {rgb}{0, .8, .2}
\definecolor{bw4}    {rgb}{0, .8, .2}
\definecolor{rw4}    {rgb}{.3, .7, .3}
\definecolor{knob3}  {rgb}{0.5, 0, 1}
\definecolor{bw3}    {rgb}{0.5, 0, 1}
\definecolor{rw3}    {rgb}{.5, .3, 1}
\definecolor{knob2}  {rgb}{0, 0, 1}
\definecolor{bw2}    {rgb}{0, 0, 1}
\definecolor{rw2}    {rgb}{0.3, .3, .9}

\newcommand{\prefix}[1]{\textcolor{prefix}{\uline{#1}}}

\newcommand{\knob}[1]{\textcolor{knob}{\textbf{[#1]}~}}
\newcommand{\rw}[1]{\textcolor{rw}{#1}}   

\newcommand{\sampend}{\ldots}

\newcommand{\appropto}{\mathrel{\vcenter{
  \offinterlineskip\halign{\hfil$##$\cr
    \propto\cr\noalign{\kern2pt}\sim\cr\noalign{\kern-2pt}}}}}

\title{Conditioned Natural Language Generation using \textit{only} Unconditioned Language Model: An Exploration}

\author{Fan-Keng Sun \\
  MIT \\
  \texttt{fankeng@mit.edu} \\\And
  Cheng-I Lai \\
  MIT \\
  \texttt{clai24@mit.edu} \\}

\date{}

\begin{document}
\maketitle
\begin{abstract}
Transformer-based language models have shown to be very powerful for natural language generation (NLG).
However, text generation conditioned on some user inputs, such as topics or attributes, is non-trivial.
Past approach relies on either modifying the original LM architecture, re-training the LM on corpus with attirbute labels, or having separately trained `guidance models' to guide text generation in decoding. 
We argued that the above approaches are not necessary, and the original unconditioned LM is sufficient for conditioned NLG.
We evaluated our approaches by the samples' fluency and diversity with automated and human evaluation. 
\end{abstract}

\section{Introduction}
We have witnessed tremendous progresses in natural language generation (NLG) with large-scale Transformer~\cite{transformer} based language models, such as the GPT-2~\cite{gpt-2}. 
A natural question to raise beyond language modeling is, how can we have more fine-grained control over these powerful models? 
Specifically, we would desire a language model to generate text centers on some user-defined conditions\footnote{Interchangeable with \textit{attributes}}. 
We call this kind of language model a \textit{conditioned} NLG. 
One could easily imagine the commercial benefits of employing a conditioned NLG in our every day products, such as search. 
To obtain a conditioned NLG, a naiive approach would be reformulate the original language modeling objectives and train the whole model from scratch~\cite{CTRL}. 
However, doing so requires tons of labeled text with conditions, which is not always available in most applications.  
Recently, PPLM~\cite{pplm} proposed to take advantage of the pretrained GPT-2 without any further retraining of the language model itself. 
This is done by having another pretrained `guidance model', such as a Bag of Word (BOW) or an attribute classifier (AC), to guide the latent state of GPT-2 toward generating more relevant tokens. 
Witnessing the upside of this approach, we further explored \textit{simple, flexible,} and \textit{effective} approaches to conditioned NLG, that encompasses these desired qualities: 
\begin{itemize}
    \item Simple: \textit{only} an unconditioned language model is needed. Does not require any additional training data~\cite{CTRL} or additional pretrained `guidance models'~\cite{pplm}.
    \item Flexible: the model should be able to model any combination of conditions with any weighting. This is not the case in previous work.
    \item Effective: the generated text is fluent and highly-relevant to the given attribute, by objective and human evaluation. 
\end{itemize}

We propose four methods in total.
Three of them are direct modeling of $p(x|c)$ by modification of token embedding or hidden states.
The other one models $p(x) p(c|x)$ by the consideration of next-token distribution.
We combine those four methods into one and compare our results with PPLM~\citep{pplm}.
Some examples of the text generated by our proposed method is in Table~\ref{tab:teaser}.

\begin{table}[!ht]
\begin{center}
  \footnotesize
\begin{tabularx}{\linewidth}{L}
    \hline
    \knob{} \prefix{To conclude}, To conclude, we have a very strong case for the hypothesis that the human body was originally made from an alien substance. There is nothing in biology or physics that would have predicted this. It's an obvious hypothesis, but if one is to believe the \sampend \\
    \hline
    \knob{Negative} \prefix{To conclude} this article with something about the role of the media in spreading the narrative of the \rw{mass shooting}, the writer has a few words about the media's role in spreading a \rw{false} news article that was originally published on Fox \sampend \\
    \hline
    \knob{Positive} \prefix{To conclude} To conclude this, let the following be the statement of his \rw{majesty}, as to the power of his \rw{Majesty} in his own name, his own body and his own blood, in all the places where the said act was made \sampend \\
    \hline
    \knob{Positive, Science} \prefix{To conclude}, the following article is a \rw{positive} review of the book by a \rw{respected scientific} and medical reviewer in the same journal entitled, The American Medical Association: A Review of Recent Medical Research on the Prevention of and \sampend \\
    \hline
    \knob{Negative, Science} \prefix{To conclude}: If the majority of \rw{scientific} papers are either \rw{negative} and \rw{false} or have a \rw{weak statistical evidence} for their own status, then these papers are not of sufficient quality for a high quality journal in the \rw{scientific} literature \sampend \\
    \hline
    \knob{Positive, Science, Military} \prefix{To conclude} this article, the US \rw{Navy} needs some of the \rw{best military research} in space and \rw{aerospace}; its \rw{first class} of space flight \rw{research ship} has a high capability for launching large cargo spacecraft to the outer \sampend \\
    \hline
    \knob{Negative, Science, Military} \prefix{To conclude}, the above is the main reason why the US \rw{military} has \rw{no} credibility for using the \rw{nuclear weapon} and the only alternative is \rw{nuclear} deterrence and \rw{nuclear} destruction in its own name – not for the other \sampend \\
\end{tabularx}
\end{center}
\caption{Our methods employ a pre-trained language model to generate text conditioned on any number of attributes without fine-tuning or human knowledge. In this table, we demonstrate results using our methods.
The underlined prefix is what the language model is given to generate a passage of text (e.g. \prefix{To conclude}). 
The controlled attributes are colored and bracketed (e.g. \knob{Science}) and we highlight words in the passage (e.g. \rw{false}) that are related to the attributes.
}\label{tab:teaser}
\end{table}

\section{Related Work}
All previous methods require knowledge about the attributes.
In~\citep{human-pref}, human knowledge about the attribute is provided to train a reward model which is used to train a NLG model by reinforcement learning.
On the other hand, \citep{CTRL} and \citep{control-style} fine-tuned NLG model on additional dataset with attribute label.
These methods usually create highly-relevant sentences but have the limitation of fixing the available attributes in advance.
In addition, PPLM-Discrim in \citep{pplm} also requires attribute label, but it only has to train a much smaller discriminator model.
Finally, PPLM-BoW in \citep{pplm} has to first construct a curated word list which is a list words that are highly related to a given attribute.
Although no fine-tuning is needed, obtaining a word list for any arbitrary attribute is definitely not a trivial task.

Although some previous methods~\citep{CTRL, pplm} do generate high-quality results, they still have major limitations when comparing to our methods.
Comparison of different methods are shown in Table.~\ref{model_comparison}.

\begin{table*}[!t]
\footnotesize
\begin{tabular}{r|c|c|l}
  \textbf{Model type}       & \textbf{Form of model} & \textbf{Samples} & \textbf{Example models and number of trainable params}
  \\ \hline
  Language model & \multirow{2}{2em}{$p(x)$}                     & \multirow{2}{3.5em}{Uncond.} &  \multirow{2}{12em}{GPT-2 medium: 345M} \\
  \citep{gpt-2} & & & \\
  \hline
  Fine-tuned language model & \multirow{2}{2em}{$p(x)$}                     & \multirow{2}{3.5em}{Uncond.} &  \multirow{2}{15em}{Fine-tuned GPT-2 medium: 345M} \\
  \citep{human-pref} & & & \\
  \hline
  Conditional language model & \multirow{2}{2.8em}{$p(x|c)$}      & \multirow{2}{2.5em}{Cond.}   &  \multirow{2}{8em}{CTRL: 1.6B} \\ \citep{CTRL} & & & \\
  \hline
  Plug and play language model & \multirow{2}{8.8em}{$p(x|c) \propto p(x)p(c|x)$} & \multirow{2}{2.5em}{Cond.} & PPLM-BoW: 0 (need curated word list) \\
  \citep{pplm} & & & PPLM-Discrim: $\sim$ 1K/attribute  \\
  \hline
  \multirow{2}{10em}{Our approaches} & $p(x|c)$ & \multirow{2}{2.5em}{Cond.} & Our-prefix: 0, Our-embedding: 0, Our-attention: 0\\
  \cline{2-2} \cline{4-4}
  & $p(x|c) \propto p(x)p(c|x)$  & & Our-next-token: 0  \\
\end{tabular}
\caption{Comparison of different methods for NLG, including unconditioned and conditioned language NLG. All conditioned methods are based on unconditioned models, but our methods doesn't require fine-tuning or any curated word list.}
\label{model_comparison}
\end{table*}

\section{Our Approaches}
We want to find approaches to model conditioned generation $p_{cg}(x_{t+1}|c, x_1, \dots, x_t)$ by using only a pre-trained language model $p_{lm}(x_{t+1}|x_1, \dots, x_t)$, where $x_i$ are the words and $c$ is the condition.
If there are $n$ condition, then $c = \{c_1, \dots, c_n\}$.
Here, we described our four methods to solve this problem.

\subsection{Our-prefix: Conditional prefix}
The first approach is the simplest one.
We feed a conditional sentence into GPT-2 before it generates the conditioned text.
For example, if we want a positive sentence about politics, then the conditional sentence to the GPT-2 will be "The following is a positive article about politics."
Although very naive, we found this method does actually work.
Empirically, we found that adding the word "following" greatly improves the coherence.

In \citep{recover-any}, the authors show that a pre-trained language model can be steered to recover arbitrary sentences.
Here, although $p_{cg}$ is definitely not close to $p_{lm}$, we think that prepending a well-designed prefix can make them closer.
The prefix alters the hidden states of the unconditioned language model in order to steer it closer to a conditioned one.
Formally, we assume that $p_{lm}(x_{t+1} | \text{``The following''} + $c$, x_1, \dots, x_t) \approx p_{cg}(x_{t+1} | c, x_1, \dots, x_t)$.

However, the problem with this method is that the model will be influenced by the added sentence.
For example, it will increase the probability of generating the word ``following'' or ``article'' subsequently.
We tried two ways to fix this.

First, we tried to disconnect the order relation of $x$ and $c$. 
We first feed $c$ into the language model and keep the key-value pairs of the self-attention.
Then, during the conditioned generation, the language model will start the generation without the input of $c$ but will self-attend on those key-value pairs.
The counting of position indices are also restarted from 0.
Unfortunately, this does not work out.
The model is greatly disturbed by those redundant key-value pairs and thus generate unrecognizable language.

Another straightforward way is to cut-off the special prefix after a fixed generation step.
This absolutely fixes the issue in a brute-force manner.
In \citep{pplm}, they also employed this method to avoid degeneration (i.e. model keep producing the same token). In this paper, we refer to this approach as `early stopping'.

\subsection{Our-embedding: changing token embedding}
Given $n$ conditions, we can use the tokenizer to obtain the token index $t_i, \forall i \in [1, n]$ corresponding to the conditions $c = \{c_1, \dots, c_n\}$.
Then, we add the token embedding of $t_i$ with weight $w_i$ to all token embeddings for all $i \in [1, n]$.
Finally, we re-normalize all embedding by dividing $1 + \sum_{i=1}^n w_i$.

In original GPT-2~\citep{gpt-2}, the input and output embeddings are tied.
Here, we untie them and only apply this change to the input embedding.
By doing so, every input embedding contains information about the conditions and the transformation from token space to embedding space is guided toward the conditions.
For example, the token ``military'' will gain positivity if the condition is ``positive'' and gain negativity if the condition is ``negative'' from the viewpoint of language model.
We do not change the output embedding because we conjecture that doing so will de-transformed the conditioned embedding.

Notice that in this method, the user can decide the weight for each condition.
This is the flexibility that we desired.

\subsection{Our-attention: changing self-attention key-value pairs}
Similar to the previous method, we change the self-attention key-value pairs in the language model by adding the key-value pairs of the condition token indices.
To obtain the key-value pairs corresponding to a token index $t_i$ at time step $t$, we feed a single token $t_i$ with position index $t$ into the model.
All key-value pairs are also re-normalize by dividing $1 + \sum_{i=1}^n w_i$.
To avoid degeneration, the weights are decrease inversely proportional to the number of time steps.

The idea of this method is similar to the previous one.
The main difference is that this method considers different position indices.

\subsection{Our-next-token: changing the output distribution by next-token distribution}
We have

\begin{align}
  & p_{cg}(x_{t+1} | c, x_1, \dots, x_t) \\
= & \frac{p(x_{t+1}, c | x_1, \dots, x_t)}{p(c | x_1, \dots, x_t)} \\
= & \frac{p(c | x_1, \dots, x_{t+1}) p(x_{t+1} | x_1, \dots, x_t)}{p(c | x_1, \dots, x_t)}.
\end{align}

Notice that $c, x_1, \dots, x_t$ all have known assignment, so $p(c | x_1, \dots, x_t)$ is a constant.
Also, $p(x_{t+1} | x_1, \dots, x_t)$ is essentially a language model.
Thus, we have
\begin{align}
\begin{split}
&p_{cg}(x_{t+1}|c, x_1, \dots, x_t) \\
&\propto p(c|x_1, \dots, x_{t+1}) p_{lm}(x_{t+1}|x_1, \dots, x_t).
\end{split}
\end{align}
In PPLM, $p(c|x_1, \dots, x_{t+1})$ is approximated by either a separate BOW or a linear classifier. 
In our approach, we use $p_{lm}(x_{t+2} | x_1, \dots, x_{t+1})$ to approximate $p(c|x_1, \dots, x_{t+1})$, that is:
\begin{align}
\begin{split}
&p_{cg}(x_{t+1}|c, x_1, \dots, x_t) \\
&\appropto p_{lm}(x_{t+2}=c|x_1, \dots, x_{t+1}) p_{lm}(x_{t+1}|x_1, \dots, x_t).
\label{eq:approx}
\end{split}
\end{align}
In practice, we can add a weight $w$ to control the influence of the condition:
\begin{align}
\begin{split}
&p_{cg}(x_{t+1}|c, x_1, \dots, x_t) \\
&\appropto p^w_{lm}(x_{t+2}=c|x_1, \dots, x_{t+1}) p_{lm}(x_{t+1}|x_1, \dots, x_t).
\label{eq:weighted}
\end{split}
\end{align}
If there are $n$ conditions, we take the weighted mean:
\begin{align}
\begin{split}
&p_{cg}(x_{t+1}|c, x_1, \dots, x_t) \\
&\appropto (\prod_{i=1}^n p^{w_i}_{lm}(x_{t+2}=c|x_1, \dots, x_{t+1}))^{\frac{1}{n}} \cdot \\
&p_{lm}(x_{t+1}|x_1, \dots, x_t).
\label{eq:mulitple}
\end{split}
\end{align}

In our implementation, we first use top-$K$ sampling to obtain $K$ next tokens $\{x^1_{t+1}, \dots, x^K_{t+1}\}$ and the next-token distribution.
Then, we feed $K$ new sequences $x_1, \dots, x_t + \{x^1_{t+1}, \dots, x^K_{t+1}\}$ into the model to have $p_{lm}(x_{t+2}|x_1, \dots, x_t, x^k_{t+1}), \forall k \in \{1, \dots, K\}$.
Next, we single out those probabilities corresponding to $x_{t+2} = c_i$.
Finally, we multiply two probabilities together with weight $w_i$ as in Equation~\ref{eq:weighted} to multinomially sample the next token.

\section{Experiments}
\subsection{Experimental Setup}
\paragraph{GPT-2 Language Model.} Our language model is based on the GPT-2, similar to that in~\cite{pplm}. We borrowed pretrained GPT-2 and PPLM models and their implementations from HuggingFace~\footnote{\href{https://huggingface.co/}{https://huggingface.co/}}. In their implementation, GPT-2 is GPT-2 medium. 

\paragraph{Hyperparameters.} The hyperparameter used in this work is detailed in Table~\ref{tab:hyperparams}.
\begin{table}
  \begin{center}
\begin{tabularx}{\linewidth}{@{}l|L@{}}
\hline
Method & Hyperparameters \\
\hline
Ours & K=12, embed-weights=0.04, attention-weights=0.02, condition-weights=0.20, early-stopping=3 \\
 \hline
PPLM-BoW & gamma=1.5, num-iterations=3, stepsize=0.03, window-length=5, kl-scale 0.01, gm-scale 0.99\\
  \hline
PPLM-Discrim & gamma=1.0 num-iterations=10 stepsize=0.04 kl-scale=0.01 gm-scale=0.95\\
\hline
\end{tabularx}
\end{center}
\caption{The full set of hyperparameters used in our work. Note that we did not perform any hyperparameter tuning.}\label{tab:hyperparams}
\end{table}
Due to time limit, we did not perform a hyperparameter sweep for our model. As described in the Appendix of~\cite{pplm}, careful hyperparameter search is vital for its generation quality, and we would imagine that our approach works much better with hyperparameter tuning.  
We directly used the hyperparameters for PPLM specified in their github repo~\footnote{\href{https://github.com/uber-research/PPLM}{https://github.com/uber-research/PPLM}}.

\paragraph{Automated Evaluation.} We evaluated the generated text by its fluency (perplexity) and diversity (Dist-1, Dist-2, Dist-3), as in ~\cite{pplm}. In our implementation, perplexity is measured by a separately pre-trained language model (GPT-2 samll); diversity is measured by the percentages of unique n-grams (1-2-3-grams)~\cite{li2015diversity}.

\paragraph{External Sentiment Classifier.} For our sentiment modeling experiments, we adopted a pre-trained tokenizer, word2vec and sentiment classifier from Twitter~\footnote{\href{https://www.kaggle.com/paoloripamonti/twitter-sentiment-analysis}{https://www.kaggle.com/paoloripamonti/twitter-sentiment-analysis}} to gauge the effectiveness of our model. The sentiment classifier is a single-layer LSTM. 

\paragraph{Human Evaluation.} We conducted a small-scale human evaluation by asking the annotators to evaluate the text by its fluency and topic relevance, both on a scale of 1-5, with 1 being 'not readable' and 5 being 'very fluent'. 
By the time of submission, we have a total of 12 annotations submitted.
For our human evaluation, we consider the following conditions and prefix:
\begin{itemize}
    \item Condition: \{Military, Religion, Politics, Science, Legal, Space, Technology, Negative, Positive\}
    \item Prefix: \{`To conclude'\}
\end{itemize}
For each condition-prefix pair, we randomly generated 10 sentences (each of 60 tokens) and picked out 3 reasonable sentences (without \textit{degeneration issue}). We do this for every pair and for both our method and PPLM, ending up with 54 sentences.
Unlike in~\cite{pplm} where A/B testing is conducted as part of its ablation study, we do not have enough time and resources to generate statistically significant text pairs. 

\paragraph{Dataset.} Given that our proposed approach \textit{does not} require any further fine-tuning, we do not need any additional corpus to obtain the conditioned NLG. 

\subsection{Single Condition Modeling}
We generated and evaluated samples based on single condition. 
Similar to~\cite{pplm}, we consider the following conditions and prefixes:
\begin{itemize}
    \item Condition: \{Military, Religion, Politics, Science, Legal, Space, Technology\}
    \item Prefix: \{`the chicken', `the house', `the potato', `the lake', `the pizza'\}
\end{itemize}
Table~\ref{tab:odd-combination} contains a few cherry-picked samples generated by our approach. 
The results of human evaluation are shown in Table~\ref{tab:single_condition}.
From the Table, we observe the classic perplexity-diversity trade-offs seen in dialog research, that although our perplexity is lower than PPLM, we achieve higher diversity scores. 
Focusing on the human evaluation columns, we can see that our approach only lag behind PPLM a little in both attribute relevance and fluency, and this is without any hyperparameter search. 
This suggests not only our approach is effective but it is certainly possible to generated conditoined natural language using only unconditioned lanugage models.  
\begin{table*}[]
\resizebox{\textwidth}{!}{
    \centering
    \begin{tabular}{@{}l | c | c | c c c c | c c c@{}}
    \toprule
         Topic & Method & Attribute relevance \% ($\uparrow$ better) & Perplexity & Dist-1 & Dist-2 & Dist-3 & Fluency ($\uparrow$ better) \\
        & & (human) &($\downarrow$ better)  & ($\uparrow$ better) & ($\uparrow$ better) & ($\uparrow$ better) &  (human) \\
                   \midrule

        \multirow{3}{*}{Military} 
        & Ours & - & 22.954 & 0.610 & 0.896 & 0.968 & - \\
        & Ours (w/ the following) & \textbf{4.167} & 22.797 & 0.597 & 0.883 & 0.964 & \textbf{3.81} \\
        & PPLM-BOW & 2.694 & 12.302 & 0.65 & 0.876 & 0.9192 & 3.472 \\
        \hline
        \multirow{3}{*}{Religion} 
        & Ours & - & 21.227 & 0.573 & 0.869 & 0.957 & - \\
        & Ours (w/ the following) & 1.472 & 20.184 & 0.552 & 0.845 & 0.941 & 3.111  \\
        & PPLM-BOW & \textbf{1.611} & 12.204 & 0.533 & 0.725 & 0.780 & \textbf{3.583} \\
        \hline
        \multirow{3}{*}{Politics} 
        & Ours & - & 21.679 & 0.581 & 0.866 & 0.949 & - \\
        & Ours (w/ the following) & 3.250 & 20.055 & 0.555 & 0.844 & 0.940 & 3.139 \\
        & PPLM-BOW & \textbf{3.278} & 12.524 & 0.660 & 0.891 & 0.935 & \textbf{3.611} \\
        \hline
        \multirow{3}{*}{Science} 
        & Ours & - & 22.645 & 0.596 & 0.887 & 0.967 & -\\
        & Ours (w/ the following) & 2.806 & 21.643 & 0.582 &0.874 & 0.958 & 3.472 \\
        & PPLM-BOW & \textbf{4.028} & 13.508 & 0.640 & 0.873 & 0.92 & \textbf{3.778} \\
        \hline
        \multirow{3}{*}{Legal} 
        & Ours & - & 22.457 & 0.598 & 0.891 & 0.968 & - \\
        & Ours (w/ the following) & 3.278 & 21.397 & 0.579 & 0.868 & 0.956 & 3.694 \\
        & PPLM-BOW & \textbf{3.528} & 12.401 & 0.662 & 0.888 & 0.930 & \textbf{4.028} \\
        \hline
        \multirow{3}{*}{Space} 
        & Ours & - & 22.529 & 0.582 & 0.881 & 0.965 & - \\
        & Ours (w/ the following) & 2.333 & 21.053 & 0.571 & 0.859 & 0.952 & 3.583 \\
        & PPLM-BOW & \textbf{3.167} & 12.101 & 0.540 & 0.728 & 0.770 & \textbf{3.639} \\
        \hline
        \multirow{3}{*}{Technology} 
        & Ours & - & 23.303 & 0.596 & 0.887 & 0.967 & - \\
        & Ours (w/ the following) & 2.861 & 23.507 & 0.578 & 0.871 & 0.957 & 3.250 \\
        & PPLM-BOW & \textbf{3.194} & 12.489 & 0.61 & 0.820 & 0.860 & \textbf{3.750} \\
        \hhline{========}
        \multirow{3}{*}{Average} 
        & Ours & - &  22.399 & 0.591 & 0.882 & 0.963 & - \\
        & Ours (w/ the following) & 2.881 & 21.520 & 0.573 & 0.863 & 0.953 & 3.437 \\
        & PPLM-BOW & \textbf{3.071} & 12.504 & 0.614 & 0.829 & 0.874 & \textbf{3.694} \\
        \hline
    \end{tabular}
}
\caption{Single Condition Modeling: automated and human evaluation results of ours approach and PPLM-BOW. 
The conditional prefix we used here is `The following is an article about <Topic>'.
In addition, we evaluated our method \textit{with} and \textit{without} the conditional prefix.  
Results here correspond to the average over all samples in each topic: <Military>, <Religion>, <Politics>, <Science>, <Legal>, <Space>, <Technology>.
20 samples are generated for each topic.
Attribute relevance and fluency is rated on a scale of 1-5. 
Perplexity implies fluency, which is computed based on an external LM \citep{gpt} different from the base LM. 
Dist-1,2,3 implies diversity, which is the percentage of unique n-grams in the samples.
}\label{tab:single_condition}
\end{table*}

\begin{table*}[]
\begin{center}
\footnotesize
\resizebox{\textwidth}{!}{
    \begin{tabular}{@{}l | c | c c | c c c c c@{}}
    \toprule
         Topic & Method & Sentiment Acc. (\%) & Sentiment Acc. (\%) & Perplexity & Dist-1 & Dist-2 & Dist-3 & {Human Evaluation}\\
           && (human)  & (external classifer) & ($\downarrow$ better)  & ($\uparrow$ better) & ($\uparrow$ better) & ($\uparrow$ better) &  Fluency ($\uparrow$ better) \\
          \midrule
          \multirow{3}{*}{Positive} 
          & Ours  & - & 60 & 26.506 & 0.590 & 0.895 & 0.972 & - \\
          & Ours (w/ the following) & 3.417 & 72 & 26.055	& 0.570 & 0.879 & 0.964 & \textbf{3.222} \\
          & PPLM-Discrim & \textbf{3.778} & 82.5 & 18.960 & 0.678 & 0.910 & 0.955 & 3.083 \\
          \hline
          \multirow{3}{*}{Negative} 
          & Ours  & - & 52 & 26.427 & 0.592 & 0.901 & 0.976 & - \\
          & Ours (w/ the following) & 2.639 & 37 & 24.820 & 0.582 & 0.886 & 0.966 & 2.944 \\
          & PPLM-Discrim & \textbf{3.361} & 70 & 12.781 & 0.638 & 0.889 & 0.940 & \textbf{3.306} \\
          \hhline{=========}
          \multirow{3}{*}{Average} 
          & Ours  & - & 56 & 26.467 & 0.591 & 0.898 & 0.974 & - \\
          & Ours (w/ the following) & 3.028 & 54.5 & 25.438	& 0.576 & 0.883 & 0.965 & 3.083 \\
          & PPLM-Discrim & \textbf{3.569} & 76.25 & 15.871 & 0.658 & 0.898 & 0.948 & \textbf{3.194} \\
          \hline
    \end{tabular}
}
\end{center}
\caption{
    Sentiment Modeling: similar analysis as in Table~\ref{tab:single_condition} is presented here. 
    Here, the conditions are <Positive> and <Negative>. 
    In addition to the metrics described in Table~\ref{tab:single_condition}, the samples are evaluated by a pretrained sentiment classifier.
    }\label{tab:sentiment}
\end{table*}

\subsection{Sentiment Modeling}
We generated and evaluated samples based on sentiments. We consider the following conditions and prefixes: 
\begin{itemize}
    \item Condition: \{Positive, Negative\}
    \item Prefix: \{`the chicken', `the house', `the potato', `the lake', `the pizza'\}
\end{itemize}
Table~\ref{tab:sentiment} contains the results of sentiment modeling. 
We can see that PPLM has better sentiment modeling in terms of human and objective evaluations. 
We suspect the reason is that the PPLM-Discrim model is fine-tuned, and its latent space is updated 10 times (num-iteration) for each generated sample, and therefore the quality is much better. 
This suggests a future extension of our approach is to also have iterative updates. 

\subsection{Multiple Conditions Modeling}
The flexibility of our approach also ensures that we can have more than 2 conditions at the same time, see Table~\ref{tab:teaser}.
Compare this to PPLM, where the conditions are pre-determined and can not be modified after the `guidance models' are trained~\cite{pplm}.

\begin{table*}[!ht]
\begin{center}
  \footnotesize
\begin{tabularx}{\linewidth}{L}
    \hline
    \knob{Military} \prefix{The chicken} wing is the most famous \rw{weapon} of the Korean \rw{military} as one of its main \rw{war-fighting} aids, so the name is usually translated into Korean as ``the chicken wing.'' \sampend \\
    \hline
    \knob{Religion} \prefix{The chicken}, which is not necessarily the most \rw{religious bird}, doesn't really enjoy eating it. It seems to like eating eggs. It actually enjoys some sort of cheese that is part of the shell. It will eat \sampend \\
    \hline
    \knob{Politics} \prefix{The chicken} of \rw{politics} is not the \rw{politician} but the \rw{political} process as embodied by the electorate. It is a \rw{political} process that is at odds with the \rw{democratic} principles on which the country was founded. \\
    \hline
    \hline
    \hline
    \knob{Military} \prefix{The horse} is an example of a \rw{military} vehicle. \rw{Military} vehicles were built to perform certain functions in particular ways and to have particular characteristics. There are lots of examples of that." And, he continues, "We have \sampend \\
    \hline
    \knob{Religion} \prefix{The horse}, and the \rw{religious} person, has a lot to go through: They have to show that their \rw{faith} and \rw{morals} are as strong as the horse's. So, what's my question to the \rw{atheists}? If \sampend \\
    \hline
    \knob{Politics} \prefix{The horse} racing industry's lobbying groups and its candidates for \rw{Congress} received \$1.6 million in total from the race promoters and their lobbyists, and the money has not gone to a single anti-slavery advocate \sampend \\
    \hline
    \hline
    \hline
    \knob{Military} \prefix{The pizza} in question is from a recent \rw{military} pizza that I ate out and didn't really like. It tasted like something that had never been served before. It had no char. It didn't have the sweetness that \sampend \\
    \hline
    \knob{Religion} \prefix{The pizza} box with the \rw{Bible's} signature in the center has the box itself on a shelf. A large cardboard cutout of \rw{Jesus Christ} is painted on the box. "I was really inspired by a movie \sampend \\
    \hline
    \knob{Politics} \prefix{The pizza} is in the form of a picture book about \rw{politics} in the USA. The man was a former student. Now he's running to be one of the next \rw{president}. As a part of this, \sampend \\
    \hline
    \hline
    \hline
    \knob{Military} \prefix{The potato} salad is not something served at \rw{military} academies. It is not something that most people eat everyday. It is a kind of food that is consumed mostly by members 
of the \rw{military}. In order to be sure \sampend \\
    \hline
    \knob{Religion} \prefix{The potato} of \rw{religion} is that belief in a \rw{god} — and when you don't believe in something, or at least can't find evidence for it, you take it for granted. But what if you' ve learned you \sampend \\
    \hline
    \knob{Politics} \prefix{The potato} has never been banned in the \rw{US}.  It's not all bad news for the potato. The Food and Drug \rw{administration} announced Tuesday that it is ending its approval process for a cancer drug that was made \sampend \\
    \hline
    \hline
    \hline
    \knob{Military} \prefix{The lake} is named after the \rw{military} officer named John Taylor, who led the charge into the Battle of Lake Tanganyika on March 1, 1854. The area that today is known as Zamboanga \sampend \\
    \hline
    \knob{Religion} \prefix{The lake} of the same name on the west side of the mountain, which is named after the Greek \rw{goddess} of the spring, is now home to a number of \rw{historic} buildings and \rw{cultural} treasures. The city was \sampend \\
    \hline
    \knob{Politics} \prefix{The lake} may be a \rw{political} issue but there is another issue with this case, as the judge said, to take care of." A hearing to determine whether the water is protected from contamination is set for Oct \sampend \\
    \hline
    \hline
    \hline
\end{tabularx}
\end{center}
\caption{
Examples generated from a designed odd combination of condition and prefix pairs.
Each example is cherry-picked from 10 samples.
The underlined prefix is what the language model is given to generate a passage of text (e.g. \prefix{To conclude}). 
The controlled attributes are colored and bracketed (e.g. \knob{Science}) and we highlight words in the passage (e.g. \rw{false}) that are related to the attributes.
Even with the odd combination, our method is still able to generate fluent samples respecting to both the attribute and prefix, though some samples are not really sensible.
}\label{tab:odd-combination}
\end{table*}

\section{Known Issues}
We think that without iterative steps (like PPLM), it is difficult to generate high-quality results.
In other words, it is difficult to produce high-quality results with only one-step update.
Also, some attributes are much difficult to obey, so it should require more updating steps.
Additionally, directly adding token embedding and self-attention key-value pairs greatly increases the perplexity.
Finally, sometimes degeneration is still observed.
This may be due to the fact of adding of token embeddings, key-values pairs, and changing of output distribution by next-token distribution.

\section{Conclusion}
Past approaches to conditioned NLG still fall short in several ways.
With that in mind, we took inspiration from recent work~\cite{pplm} and proposed four methods for conditioned NLG that is simple, flexible and effective, such that only the original base LM is needed.
We displayed a few samples for single and multiple conditions NLG.
Experiments are conducted for single condition modeling and sentiment modeling, and these samples are evaluated based on their fluency and diversity.
A note on the inadequacy of our appraoches is appended at the end. 
\bibliography{anthology,acl2020}

\begin{thebibliography}{9}
\expandafter\ifx\csname natexlab\endcsname\relax\def\natexlab#1{#1}\fi

\bibitem[{Dathathri et~al.(2020)Dathathri, Madotto, Lan, Hung, Frank, Molino,
  Yosinski, and Liu}]{pplm}
Sumanth Dathathri, Andrea Madotto, Janice Lan, Jane Hung, Eric Frank, Piero
  Molino, Jason Yosinski, and Rosanne Liu. 2020.
\newblock Plug and play language models: A simple approach to controlled text
  generation.
\newblock In \emph{International Conference on Learning Representations}.

\bibitem[{Ficler and Goldberg(2017)}]{control-style}
Jessica Ficler and Yoav Goldberg. 2017.
\newblock Controlling linguistic style aspects in neural language generation.
\newblock In \emph{Proceedings of the Workshop on Stylistic Variation}, pages
  94--104.

\bibitem[{Keskar et~al.(2019)Keskar, McCann, Varshney, Xiong, and
  Socher}]{CTRL}
Nitish~Shirish Keskar, Bryan McCann, Lav Varshney, Caiming Xiong, and Richard
  Socher. 2019.
\newblock Ctrl - a conditional transformer language model for controllable
  generation.
\newblock In \emph{arXiv preprint arXiv:1909}.

\bibitem[{Li et~al.(2015)Li, Galley, Brockett, Gao, and
  Dolan}]{li2015diversity}
Jiwei Li, Michel Galley, Chris Brockett, Jianfeng Gao, and Bill Dolan. 2015.
\newblock A diversity-promoting objective function for neural conversation
  models.
\newblock \emph{arXiv preprint arXiv:1510.03055}.

\bibitem[{Radford et~al.(2018)Radford, Narasimhan, Salimans, and
  Sutskever}]{gpt}
Alec Radford, Karthik Narasimhan, Tim Salimans, and Ilya Sutskever. 2018.
\newblock \href
  {https://s3-us-west-2.amazonaws.com/openai-assets/research-covers/language-unsupervised/language\_understanding\_paper.pdf}
  {Improving language understanding by generative pre-training}.

\bibitem[{Radford et~al.(2019)Radford, Wu, Child, Luan, Amodei, and
  Sutskever}]{gpt-2}
Alec Radford, Jeffrey Wu, Rewon Child, David Luan, Dario Amodei, and Ilya
  Sutskever. 2019.
\newblock Language models are unsupervised multitask learners.
\newblock In \emph{OpenAI Blog}.

\bibitem[{Subramani et~al.(2019)Subramani, Bowman, and Cho}]{recover-any}
Nishant Subramani, Sam Bowman, and Kyunghyun Cho. 2019.
\newblock Can unconditional language models recover arbitrary sentences?
\newblock In \emph{arXiv preprint arXiv:1907.04944}.

\bibitem[{Vaswani et~al.(2017)Vaswani, Shazeer, Parmar, Uszkoreit, Jones,
  Gomez, Kaiser, and Polosukhin}]{transformer}
Ashish Vaswani, Noam Shazeer, Niki Parmar, Jakob Uszkoreit, Llion Jones,
  Aidan~N Gomez, Łukasz Kaiser, and Illia Polosukhin. 2017.
\newblock Attention is all you need.
\newblock In \emph{Advances in Neural Information Processing Systems}, pages
  6000--6010.

\bibitem[{Ziegler et~al.(2019)Ziegler, Stiennon, Wu, Brown, Radford, Amodei,
  Christiano, and Irving}]{human-pref}
Daniel~M. Ziegler, Nisan Stiennon, Jeffrey Wu, Tom~B. Brown, Alec Radford,
  Dario Amodei, Paul Christiano, and Geoffrey Irving. 2019.
\newblock \href {https://arxiv.org/abs/1909.08593} {Fine-tuning language models
  from human preferences}.
\newblock In \emph{arXiv preprint arXiv:1909.08593}.

\end{thebibliography}
\bibliographystyle{acl_natbib}

\end{document}